\documentclass[conference]{IEEEtran}

\ifCLASSINFOpdf
\else
\fi

\usepackage{graphicx}
\graphicspath{ {im/} }

\begin{document}

\title{Outlier absorbing based on a Bayesian approach}

\author{\IEEEauthorblockN{Parsa Bagherzadeh}
\IEEEauthorblockA{Department of Computer Engineering\\Ferdowsi Unievrsity of Mashhad\\
Mashhad, Iran +989358435990\\
Email: parsa.bagherzadeh@stu.um.ac.ir}
\and
\IEEEauthorblockN{Hadi Sadoghi Yazdi}
\IEEEauthorblockA{Department of Computer Engineering\\Ferdowsi Unievrsity of Mashhad\\
Mashhad, Iran \\
Email: h-sadoghi@um.ac.ir}
}

\maketitle

\begin{abstract}
The presence of outliers is prevalent in machine learning applications and may produce misleading results. In this paper a new method for dealing with outliers and anomal samples is proposed. To overcome the outlier issue, the proposed method combines the global and local views of the samples. By combination of these views, our algorithm performs in a robust manner. The experimental results show the capabilities of the proposed method.
\end{abstract}


\IEEEpeerreviewmaketitle

\section{Introduction}
 Data quality is one of greatest concerns in data mining and machine learning. Most of machine learning methods perform inaccurately or produce misleading results when data suffers from lack of quality. Limitation of measuring instruments, human error in the data equation process may lower data quality. In some cases, the value of a feature may be missing. In other cases, the data may be contaminated by external sources and not indicating their real value \cite{Vipin}. \\

One of main issues is the context of data quality is the presence of outliers. Outliers are instances which have considerable difference with the majority of instances. Another outlier definition from \cite{Barn} is: A sample (or subset of samples) which appears to be inconsistent with the rest of that data set. An outlier may also be surprising veridical 
data, a sample belonging to class $\omega_1$ but actually positioned inside class $\omega_2$ so the true (veridical) classification of the sample is surprising to the observer (this type of outlier is also called label noise). \\

The presence of outliers may cause potential problems in both supervised and unsupervised learning. The most significant consequence of label noise is degradation of classification performance \cite{Mich}, \cite{Bi}. For example it is shown that only $5\%$ of outliers can highly deviate the decision boundaries. In \cite{Zhang} SVMs, ridge regression, and logistic regressions are tested is the presence of outliers. The experiments show that the results are highly affected by outliers for all three methods. \\

Moreover, outliers may cause over-fitting on training data. The presence of outliers also increases the required number of instances for learning, as well as the complexity of models \cite{Fr}. In \cite{Brod} it is shown that the removal of outliers reduces the number of support vectors. Non-robust classifier methods produce models which are skewed when outliers are left in.  An example of a data set suffering from outliers is illustrated in Fig. 1. The outliers are indicated by small red circles around.\\

\begin{table}
\begin{center}
\begin{tabular}{c}
\includegraphics[scale=0.35]{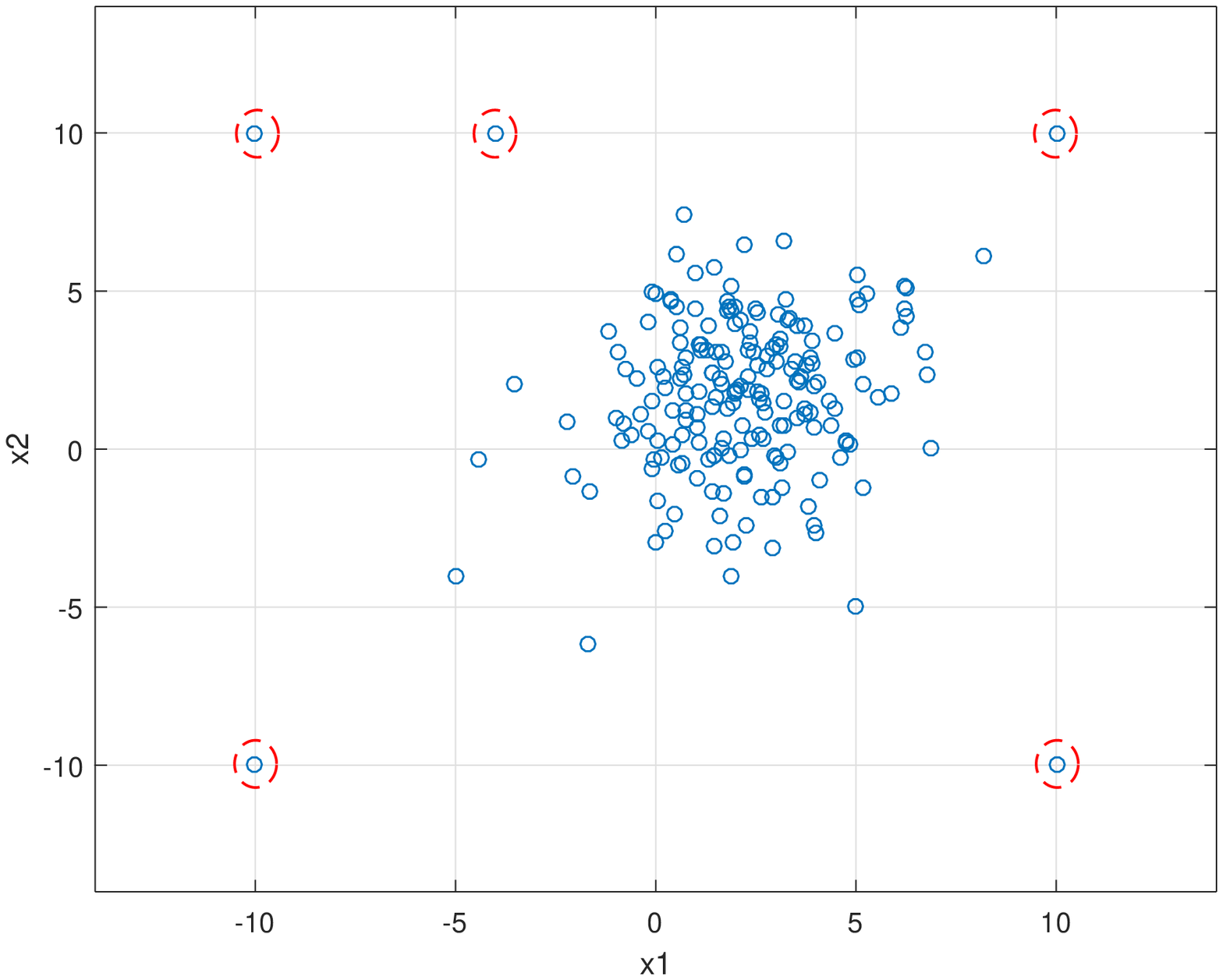}\\
\textbf{Fig 1.} A data set suffering form outliers
\end{tabular}
\end{center}
\end{table}

There are various methods for detection of outliers \cite{Hod}. In this paper we focus on proximity-based techniques including k-NN-based methods. These methods are simple to implement and make no prior assumptions about the data distribution model. \\

Ramaswamy et al. introduced an optimized k-NN to produce a ranked list of potential outliers \cite{Rams}. A sample is an outlier if no more than $n-1$ other points in the data set have a higher $D_m$ (distance to mth neighbor) where $m$ is a user-specified parameter. Since most of k-NN based approaches are susceptible to the computational growth several techniques were proposed for speeding the k-NN algorithm such as partitioning the data into cells. If any cell and its adjacent cells contains more than $k $ points, then the points in the cell are probably lied in a dense area of the distribution so the points contained are unlikely to be outliers. \\

Another proximity-based variant is the graph connectivity method. Shekhar et al. introduced an approach for traffic monitoring which views the outlier issue from a topologically perspective \cite{Shekh}. Shekhar detects traffic monitoring stations producing sensor values which are inconsistent with stations in the connected neighborhood. A station is an outlier if the difference between its sensor value and the average sensor value of its topological neighbors differs significantly from the mean difference between all nodes and their topological neighbors. \\

Knorr and Ng (1998) introduced an efficient type 1 k-NN approach. If $m$ of the $k$ nearest neighbors (where $m < k$) lie within a specific distance threshold d then the exemplar is deemed to lie in a sufficiently dense region of the data distribution to be classified as normal. However, if there are less than $m$ neighbors inside the distance threshold then the exemplar is an outlier.\\

Several problems are accompanying with k-NN based approaches. Most of k-NN based approaches only view data locally. This approach may fail when there are batches of outliers in data set. Another approach is to have a global view of samples. In global view, samples with large distance from the distribution of samples are detected as outlier. A potential problem however is the determination of a threshold. An inappropriate threshold may lead to detection of a correct sample as outlier.\\ 
 
In this paper a new kNN-based method for dealing with outliers is proposed. The proposed method solves the problems of kNN-based method by combining the local and global views of samples. \\  

The rest of this paper is organized as follows: Section 2 presents our method for outlier absorbing. In Section 3 the experimental results are investigated and finaly section 4 give the concluding remarks. 

\section{The proposed method}\label{section3}
In this section our proposed method for outlier absorbing is presented. The proposed method combines the local and global information of sample to achieve more robust results. 
\subsection{Notations}
Let $\Omega_x$ be the state space of traning samples, $X$.  In other words

\begin{equation}\label{eq2}
\Omega_x=\{x_1,x_2,\dots,x_n\}
\end{equation}

If a sample $x$ be noisy, it is desired to estimate $\hat{x}$, as a new noise-free instance. Let $\Omega_{x,k}$ be the set of all k nearest neighborhoods and $x_i^{kNN}$ be the nearest neighbors of instance $x_i$. Also suppose $\Omega_x^{-i}$ be the set of all samples except the $i$th instance and $\Omega_{x,k}^{-i}$ be the set of all k nearest neighborhoods except the k nearest neighbors of $i$th instance. 

\subsection{Markovian-like assumption}
Markovian assumption holds for state spaces in which a sequence of states occurs temporally so that the probability of being in a sate at time $t$ is only given it's previous state, not all of the previous states in the sequence. In other words

\begin{equation}
f(x_t|x_{t-1},x_{t-2},...)=f(x_t|x_{t-1})
\end{equation}\\

The main notion of Markovian property is that when a state $x_{t-1}$ explicitly contains the information of other states $\{x_{t-2},x_{t-2},...\}$, these states can be ignored. This property holds when the states have a temporal nature. We are looking for a same property when the states have a spatial nature. Similar to Markovian assumption, if a set of samples $S_1$ contains the information of another set $S_2$ it is reasonable to ignore set $S_2$. Fore example, consider two sets $\Omega_{x,k}^{-i}$ and $x_i^{kNN}$. Each instance in $x_i^{kNN}$ can be represented by its k nearest neighbors in $\Omega_{x,k}^{-i}$ thus a Markovian-like property holds for these two sets. Along with our problem formulation, we will use this Markovian-like property.

\subsection{Problem formulation}
Let $f(x|\Omega_{x,k})$ be the probability density function of $x$ given the set of all k nearest neighborhoods. A representation of $f(x|\Omega_{x,k})$ is weighted perfect sampling.

\begin{equation}\label{eq6}
f(x|\Omega_{x,k})=\frac{\sum_{i \in \Omega_{x,k}} w_i(x) \delta(x-x_i)}{\sum_{i \in \Omega_{x,k}} w_i(x)}  
\end{equation} 
where $w_i(x)$ is the weight of instance $x_i$ to be defined later. 

$\hat{x}$ should be extracted from $f(x|\Omega_{x,k})$. It can be defined as expected value of $g(x)$ over $f(x|\Omega_{x,k})$ where $g(x)$ is an arbitrary loss function. In other words

\begin{equation}\label{eq7}
\hat{z}=E\{g(x)\}=\int g(x)f(x|\Omega_{x,k}) dx. 
\end{equation} 

In the case of $g(x)=x$ and using \ref{eq6} we have:

\begin{equation}\label{eq8}
\hat{x}= \frac{ \sum_{i \in \Omega_{x,k}} w_i(x) x_i }{\sum_{i \in \Omega_{x,k}} w_i(x)}
\end{equation} 
where $x_i$ is one of kNN samples (Look at the Appendix for details). The recent equation is the representation of an instance based on its k nearest neighbors which means a representation based on a local view to samples.\\

Let $f(\Omega_x|\Omega_{x,k})$ be the PDF of $\Omega_x$ given the set of all k nearest neighborhoods and suppose that $w_i(x)$, the weight of instance $x_i$ be defined as:
\begin{equation}
w_i(x)=\frac{f(\Omega_x|\Omega_{x,k})}{q(\Omega_x|\Omega_{x,k})}
\end{equation}

We can decompose $f(\Omega_x|\Omega_{x,k})$ as follow:

\begin{center}
\large
$f(\Omega_x|\Omega_{x,k})=\frac{f(x_i,\Omega_x^{-i},\Omega_{x,k}^{-i},x_i^{kNN})}{f(\Omega_{x,k})}$

\end{center}

\begin{equation}\label{eq11}
={\frac{f(x_i^{kNN}|x_i,\Omega_x^{-i},\Omega_{x,k}^{-i})f(x_i|\Omega_x^{-i},\Omega_{x,k}^{-i})f(\Omega_x^{-i}| \Omega_{x,k}^{-i})f(\Omega_{x,k}^{-i})}{f(\Omega_{x,k})}}
\end{equation}

If the number of samples are sufficiently large, we can assume $f(\Omega_{x,k})\simeq f(\Omega_{x,k}^{-i})$, thus

\begin{equation}\label{eq12}
f(\Omega_x|\Omega_{x,k})\propto f(x_i^{kNN}|x_i,\Omega_x^{-i},\Omega_{x,k}^{-i})f(x_i|\Omega_x^{-i},\Omega_{x,k}^{-i})f(\Omega_x^{-i}| \Omega_{x,k}^{-i})
\end{equation}
and by Markovian-like assumption

\begin{equation}
f(\Omega_x|\Omega_{x,k})\propto f(x_i^{kNN}|\Omega_x^{-i})f(x_i|\Omega_x^{-i})f(\Omega_x^{-i}| \Omega_{x,k}^{-i})
\end{equation}

Using a slightly different decomposition for $q(\Omega_x|\Omega_{x,k})$ we can write:

\begin{center}
\large
$q(\Omega_x|\Omega_{x,k})=\frac{q(\Omega_x^{-i},x_i,\Omega_{x,k})}{q(\Omega_{x,k})}
$
\end{center}

\begin{center}
\large
$=\frac{q(x_i|\Omega_x^{-i},\Omega_{x,k})q(\Omega_x^{-i}|\Omega_{x,k})q(\Omega_{x,k})}{q(\Omega_{x,k})}$
\end{center}

\begin{equation}\label{eq13}
=q(x_i|\Omega_x^{-i},\Omega_{x,k})q(\Omega_x^{-i}|\Omega_{x,k})
\end{equation}

 By Markovian-like assumption:

\begin{equation}\label{eq14}
q(\Omega_x^{-i}|\Omega_{x,k}) \simeq q(\Omega_x^{-i}|\Omega_{x,k}^{-i})q(x_i^{kNN}|\Omega_{x,k}^{-i})
\end{equation}
(Look at the Appendix for details). 

Thus:

\begin{center}
$w_i^{itn}(x)=\frac{f(\Omega_x|\Omega_{x,k})}{q(\Omega_x|\Omega_{x,k})}$
\end{center}

\begin{equation}\label{eq15}
 \propto\frac{f(x_i^{kNN}|\Omega_x^{-i})f(x_i|\Omega_x^{-i})f(\Omega_x^{-i}| \Omega_{x,k}^{-i})}{q(x_i|\Omega_x^{-i},\Omega_{x,k})q(\Omega_x^{-i}|\Omega_{x,k}^{-i})q(x_i^{kNN}|\Omega_{x,k}^{-i})}
\end{equation}
taking 

\begin{equation}\label{eq16}
w_i^{itn-1}(x)=\frac{f(\Omega_x^{-i}| \Omega_{x,k}^{-i})}{q(\Omega_x^{-i}|\Omega_{x,k}^{-i})}
\end{equation}
yields the following recursive update equation: 

\begin{equation}\label{eq17}
w_i^{itn}(x)=w_i^{itn-1}(x)\frac{f(x_i^{kNN}|\Omega_x^{-i})f(x_i|\Omega_x^{-i})}{q(x_i|\Omega_x^{-i},\Omega_{x,k})q(x_i^{kNN}|\Omega_{x,k}^{-i})}
\end{equation}
for simplicity we assume
\begin{equation}\label{eq18}
\frac{f(x_i^{kNN}|\Omega_x^{-i})}{q(x_i|\Omega_x^{-i},\Omega_{x,k})q(x_i^{kNN}|\Omega_{x,k}^{-i})}=1
\end{equation}
which yields an update equation for weights of samples
\begin{equation}\label{eq19}
w_i^{itn}(x)=w_i^{itn-1}f(x_i|\Omega_x^{-i})
\end{equation}
where $f(x_i|\Omega_x^{-i})$ means the evaluated value of the PDF of all samples except $x_i$ at instance $x_i$, which corresponds to a global view samples for estimation of $\hat{x}$ using \ref{eq8}. A realization of  $f(z_i|\Omega_z^{-i})$ could be obtained using GMMs (Gaussian mixture models).  The steps of the proposed algorithm for label denoising is presented as follows:
\\

\textbf{Outlier absorbibg based on a Bayesian approach
}

\begin{itemize}
\item Input: Data matrix $X \in \Re^{n\times d}$ 
\item Initialization: Set the weights of all samples to $\frac{1}{n}$, where $n$ is the number of samples.
\item Step 1: Update the weights using equation \ref{eq19}. 
\item Step 2: For each instance $z$, estimate $\hat{z}$ with respect to weights of its neighbors using equation \ref{eq8} and update $z$ with the estimation.
\item Step 3: If $Div(f^{itn}(\Omega_z|\Omega_{z,k})||f^{itn-1}(\Omega_z|\Omega_{z,k}))<\epsilon$ or maximum numbers of iterations reached, then terminate, otherwise go to Step 1.
\item Output: Denoised data set
\end{itemize}

\section{Experimental results}
\subsection{Artificial Data sets}
We have applied the proposed method to two artificial data sets. First, the method is applied to a gaussian distribution. Then the non-linear distribution case is considered. 
\subsubsection{Gaussian distribution}

In order to evaluate the proposed method, different portions of outliers are added to the artificial data set. 150 instances  drawn from a Gaussian distribution (Fig. 2.a). These smaples are obtained by adding Gaussian noise to randomly selected samples. The evaluation of the proposed method is performed in the presence of outlier with different percentages including $5\%$, $10\%$, $15\%$ and $20\%$. A demonstration of the case with $10\%$ outliers is shown in Fig. 2.\\

\begin{table*}
\begin{center}
\begin{tabular}{ccc}
\includegraphics[scale=0.38]{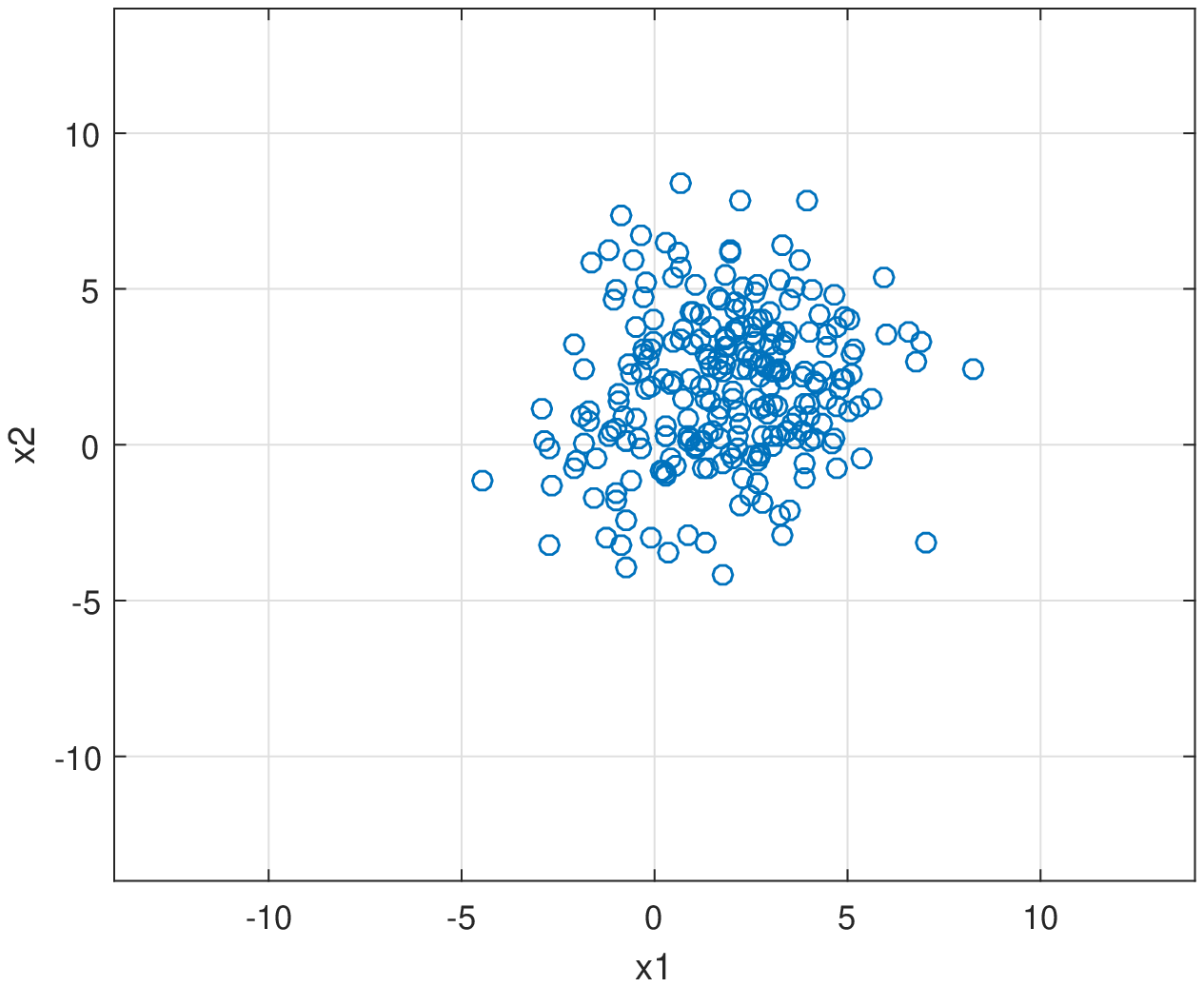}&\includegraphics[scale=0.38]{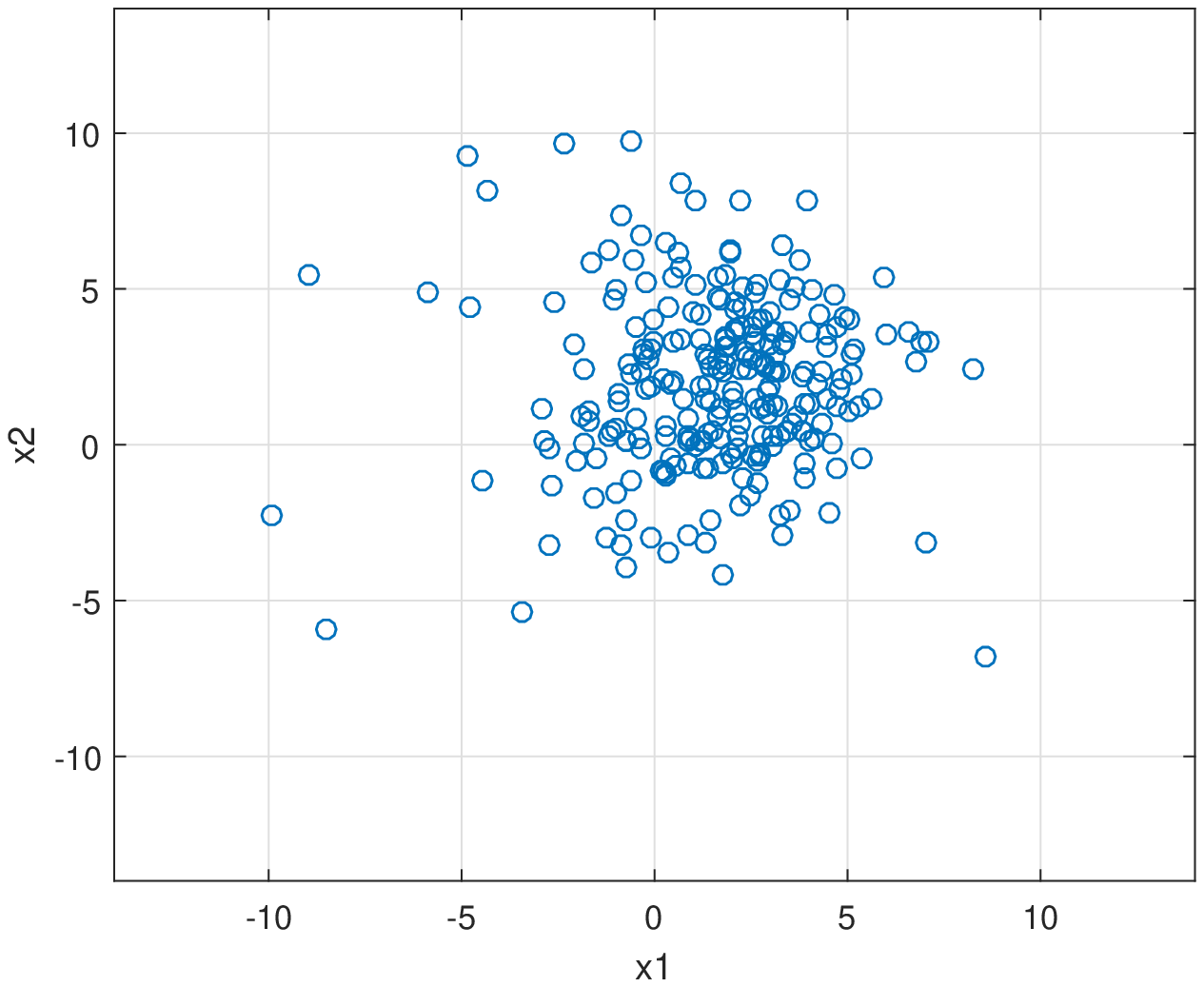}&\includegraphics[scale=0.38]{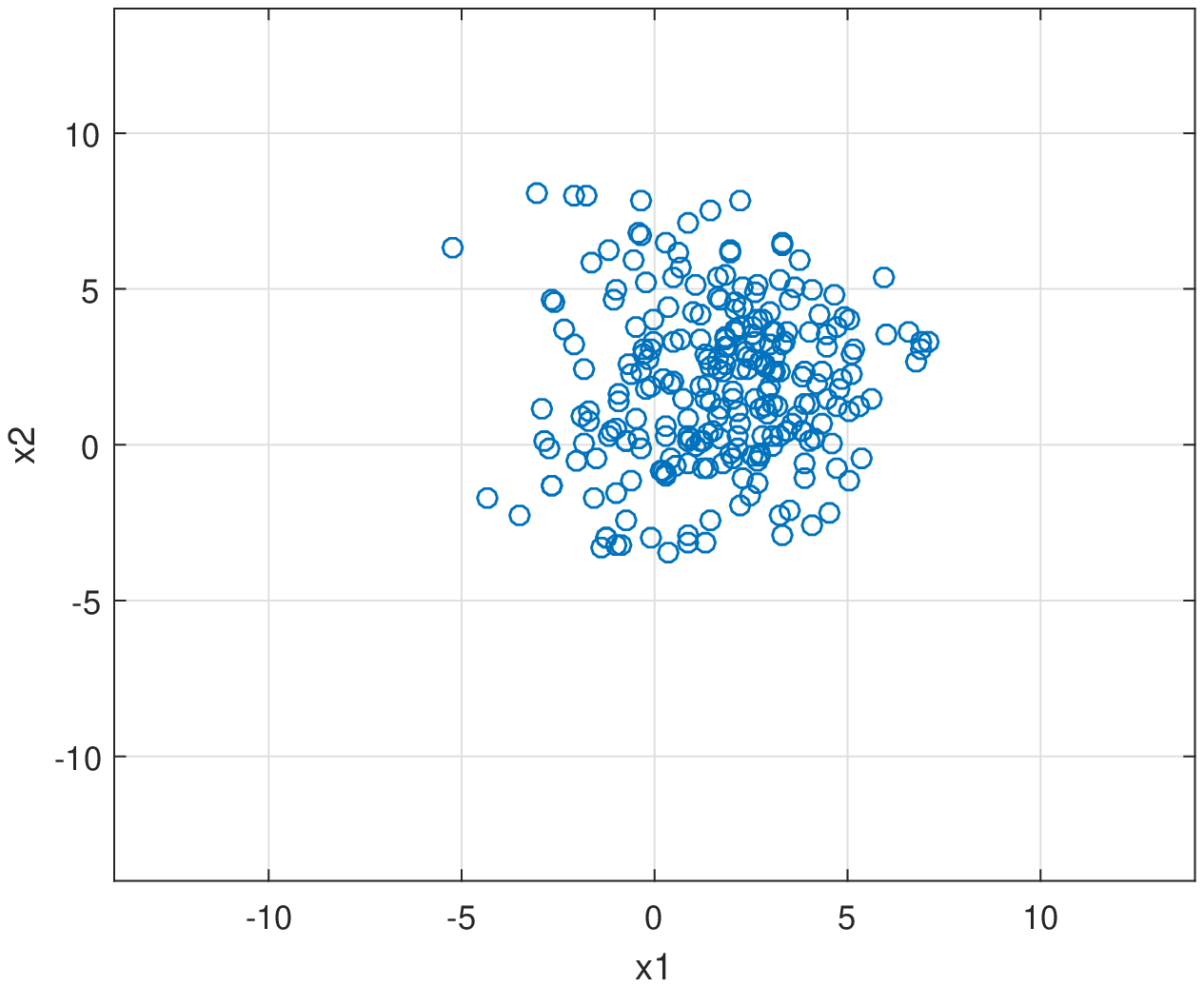}\\
a) Initial distirbution &b) The distribution with $10\%$ outliers &c) After outlier absorbing \\
\\
\multicolumn{3}{c}{\textbf{Fig. 2} Performance of the proposed method on a Gaussian distribution}
\end{tabular}

\end{center}
\end{table*}

\subsubsection{Non-linear distribution}
A challenging case in outlier detection problem is the case of non-liner data. An example of a non-liner distribution is illustrated in Fig.  3. As our experiments show, out proposed method for outlier absorbing is also robust for these types of distributions. Fig. 3 and Fig. 4 show the performance of the proposed method for the case with $10\%$ and $15\%$ outlier respectively. \\

\begin{table*}[ht]
\begin{center}
\begin{tabular}{c}
\includegraphics[scale=0.354]{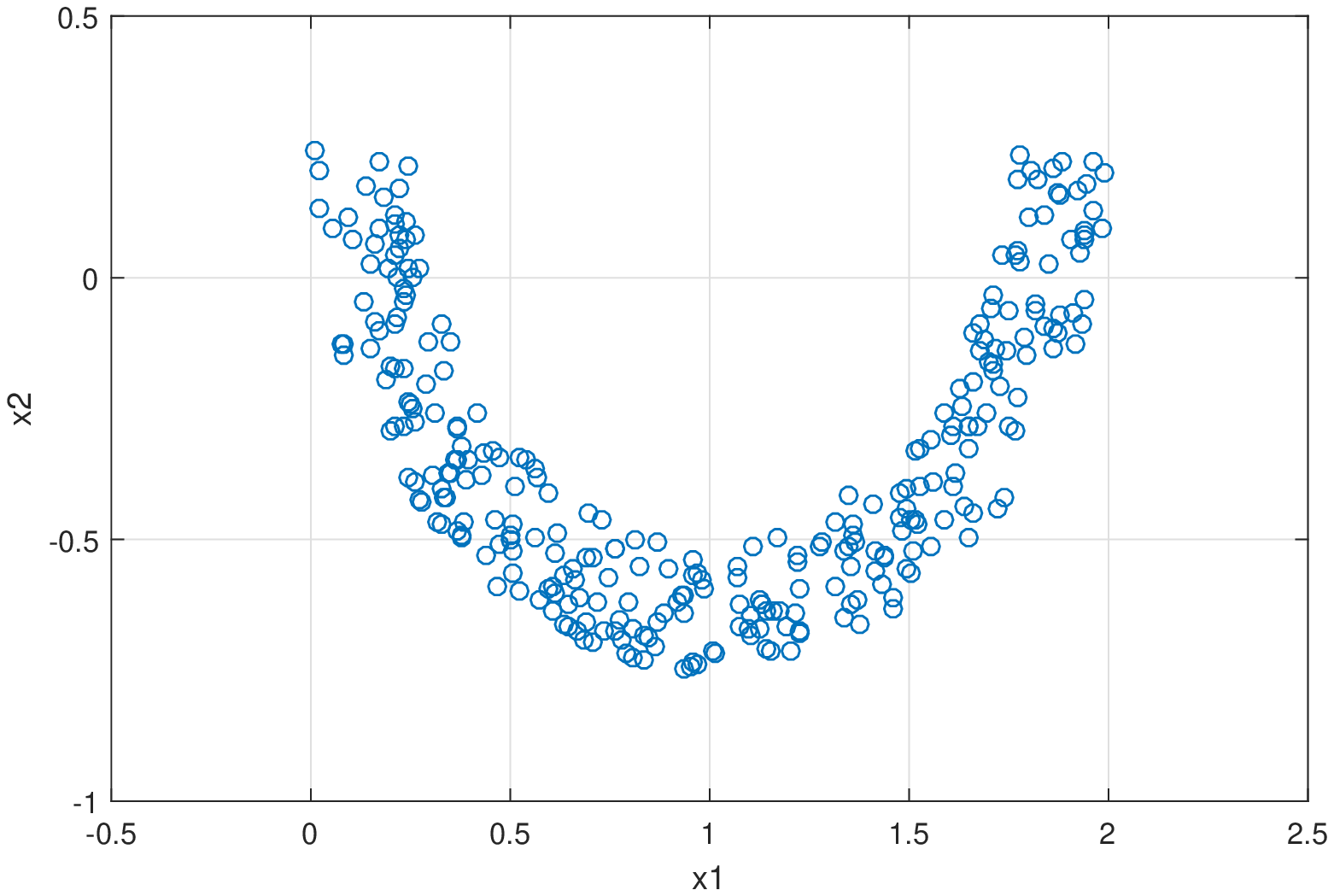}\\
\\
\textbf{Fig. 3} : A non-linear distribution for evaluation of the proposed method

\end{tabular}
\end{center}
\end{table*}

\begin{table*}[ht]
\begin{center}
\begin{tabular}{ccc}
\includegraphics[scale=0.28]{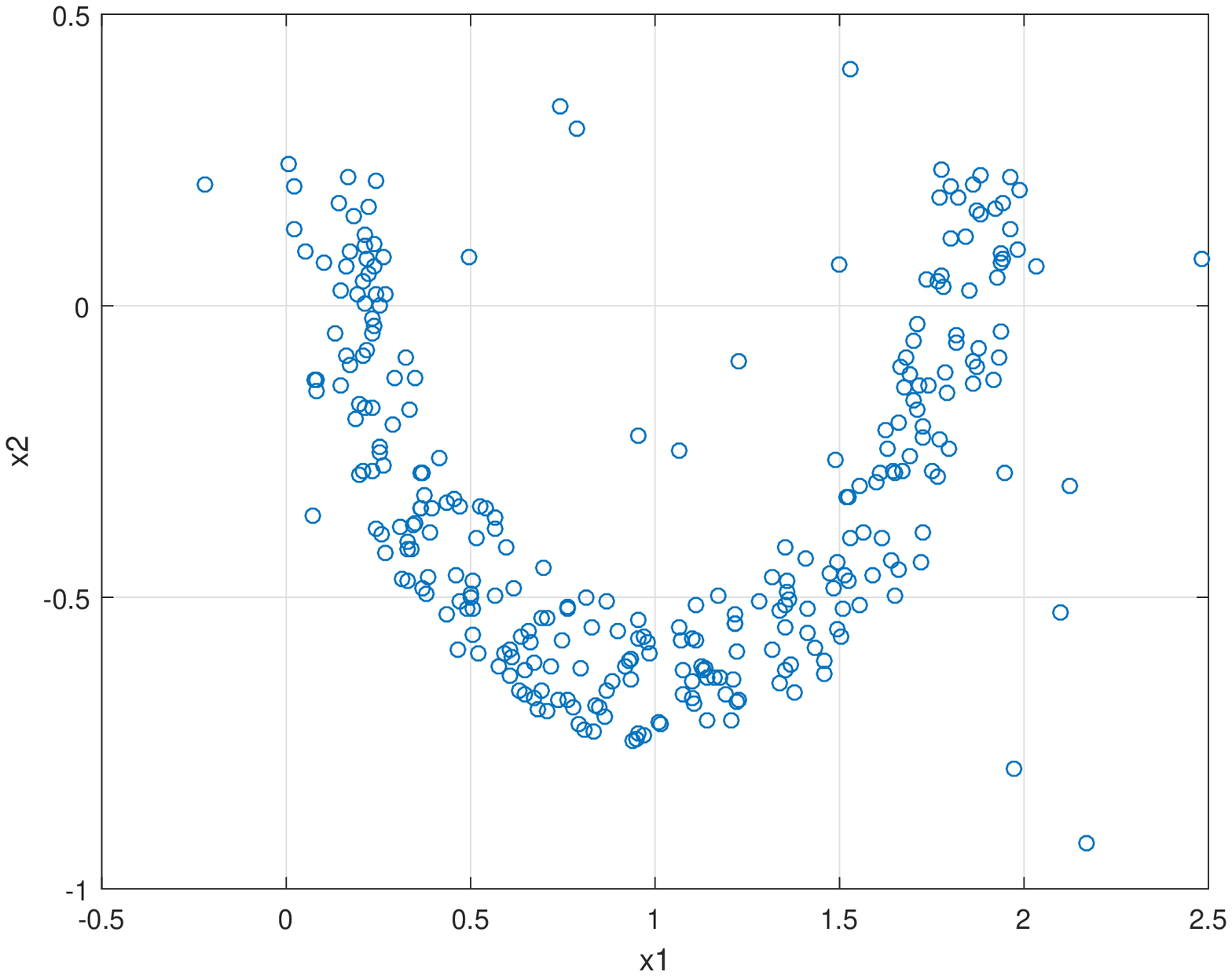}& \includegraphics[scale=0.28]{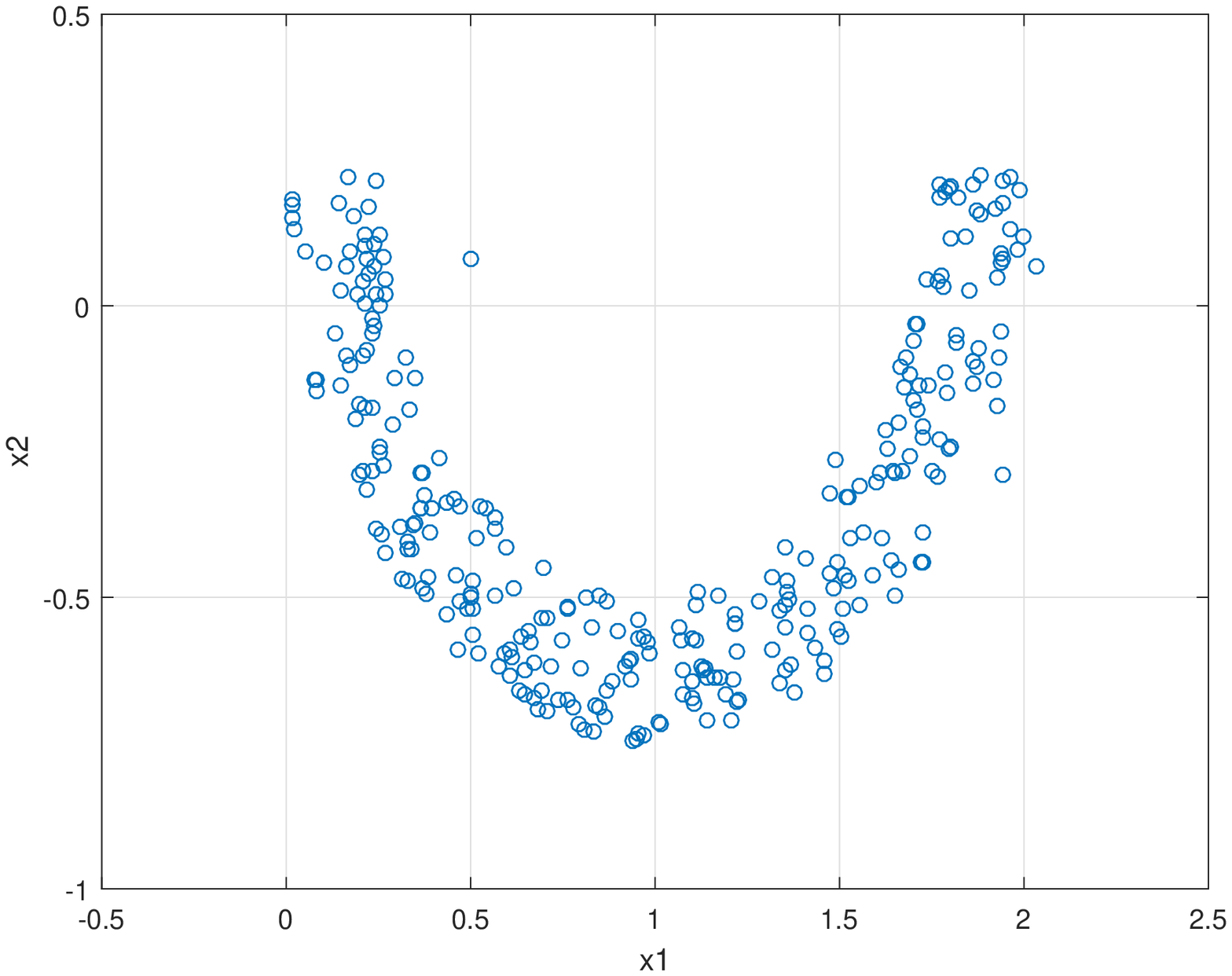}& \includegraphics[scale=0.28]{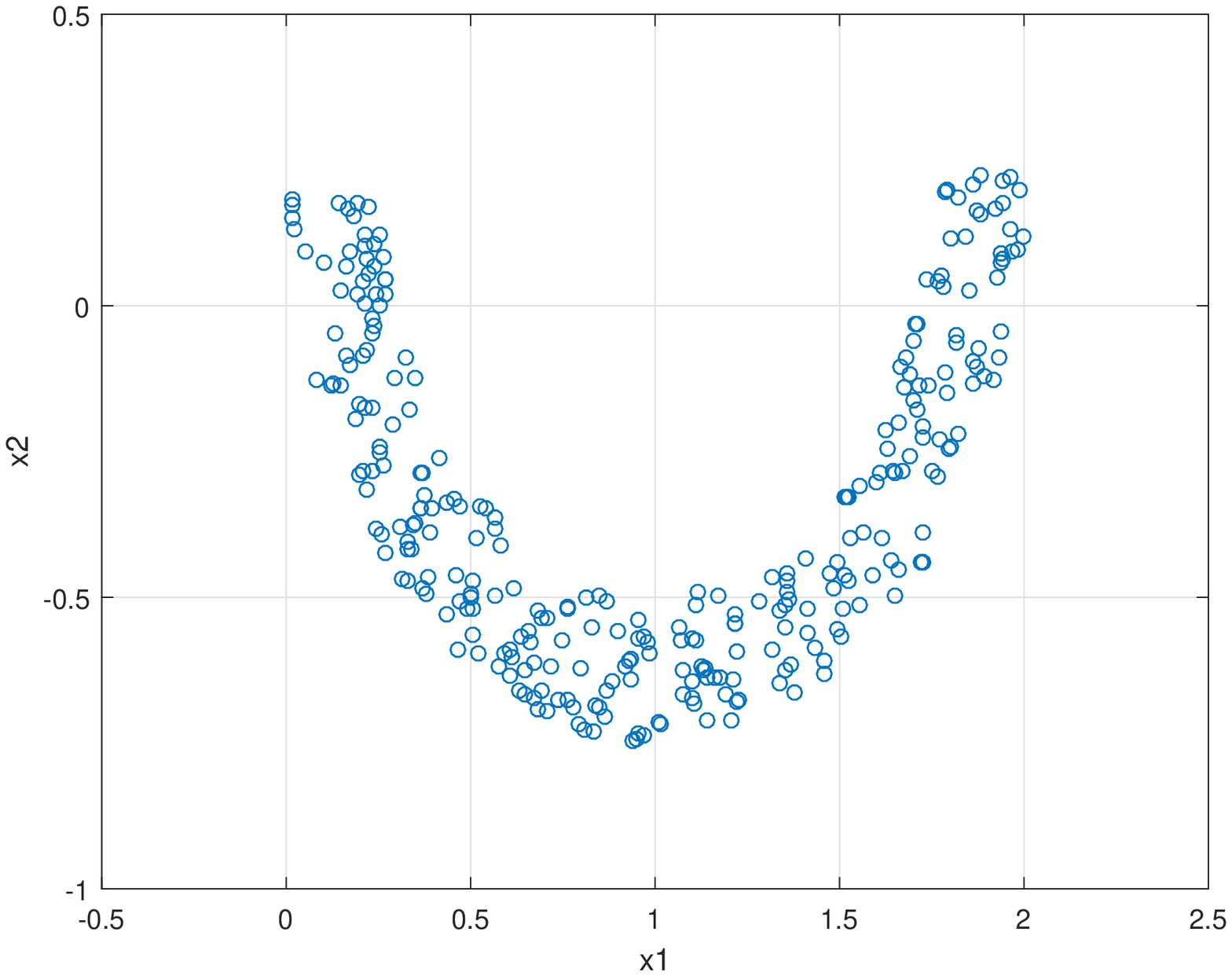}\\

a) The distribution with $10\%$ outliers & b) Iteration $\# 2$ &c) Iteration $\# 4$\\
\\
\multicolumn{3}{c}{\textbf{Fig. 4} Performance of the proposed method for non-linear distribution in the presnece of $10\%$ outliers.}
\end{tabular}
\end{center}

\end{table*}

\begin{table*}[ht]
\begin{center}
\begin{tabular}{cc}
\includegraphics[scale=0.3]{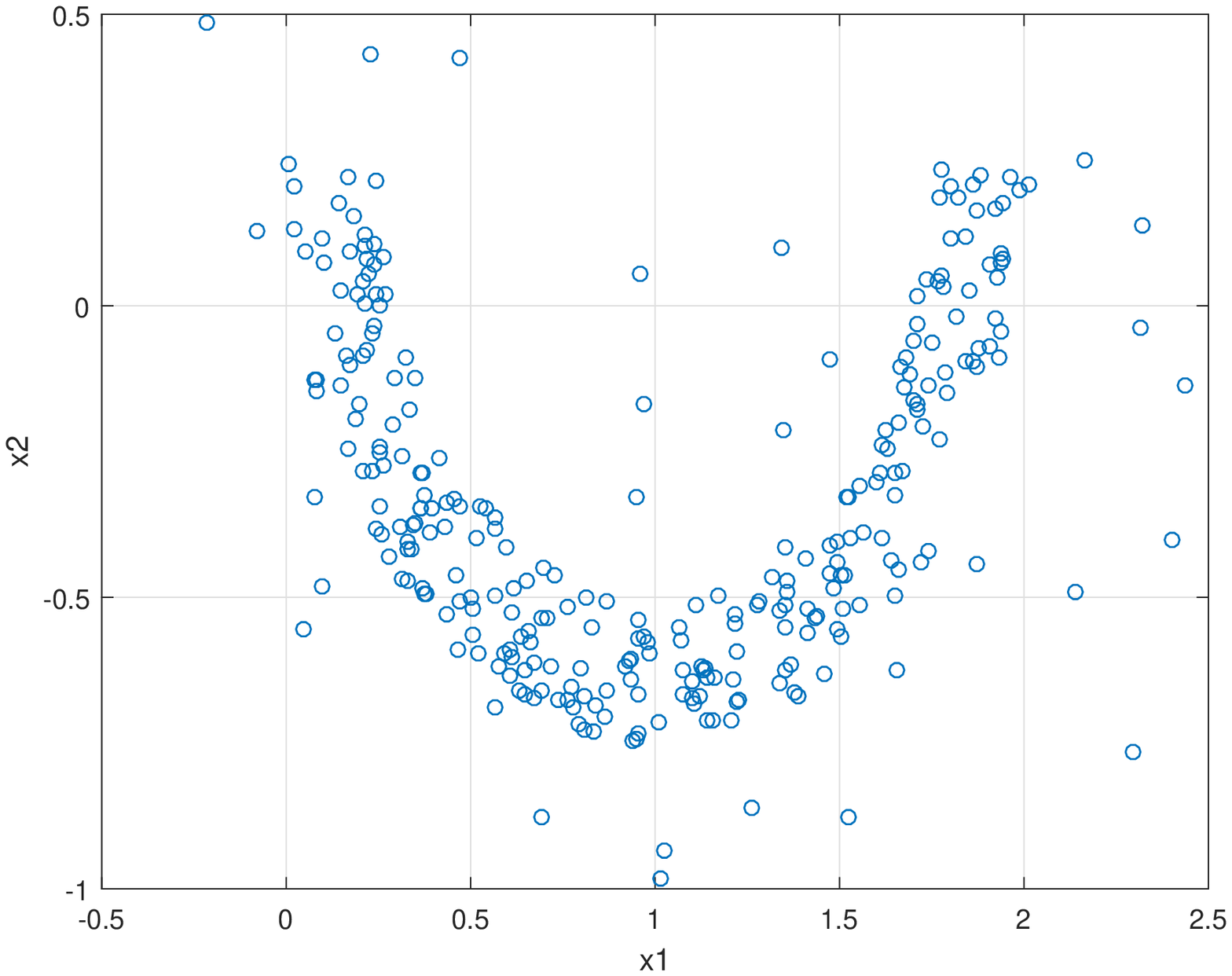}& \includegraphics[scale=0.3]{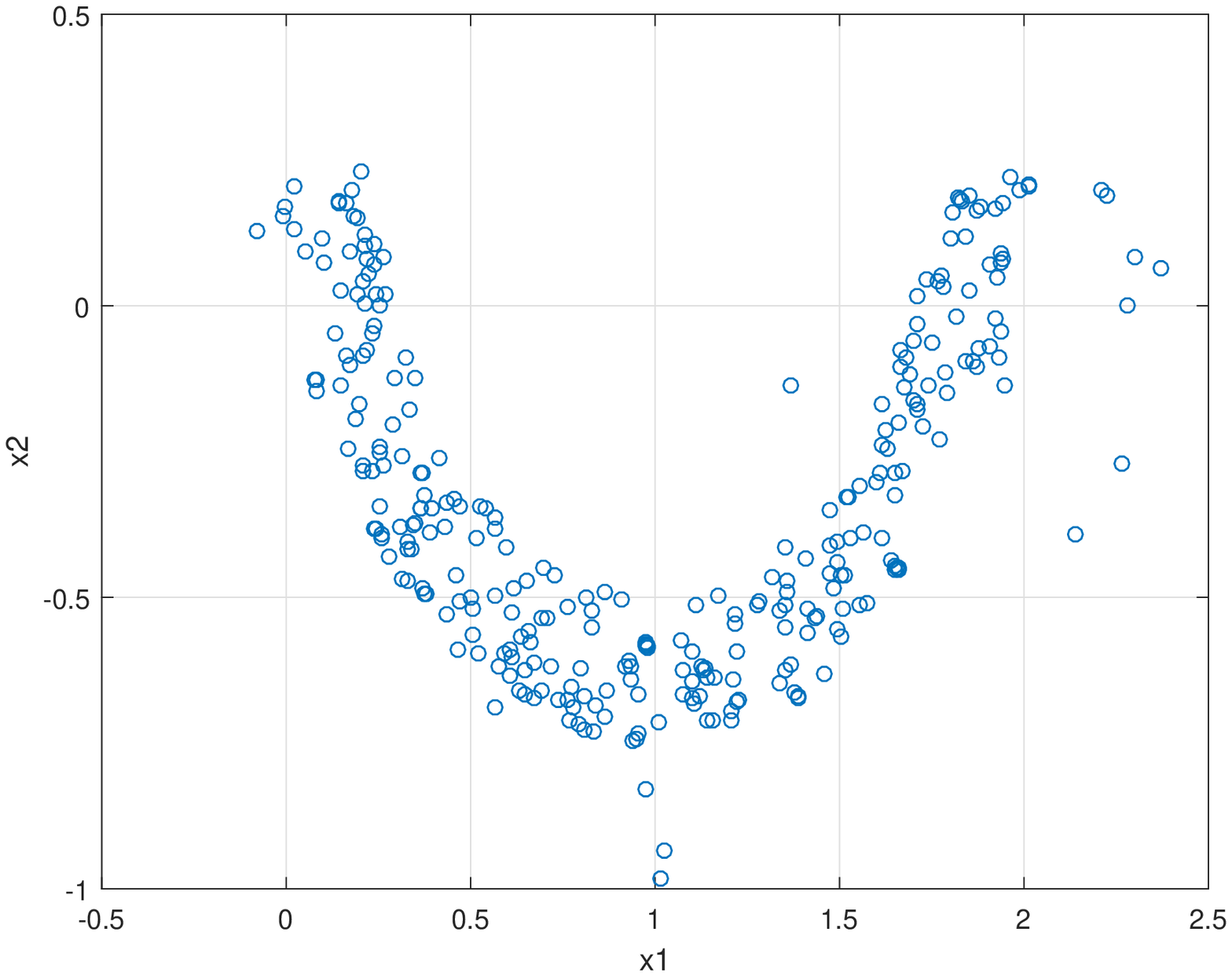}\\
a) The distribution with $15\%$ outliers  & b) Iteration $\#2$\\
\includegraphics[scale=0.3]{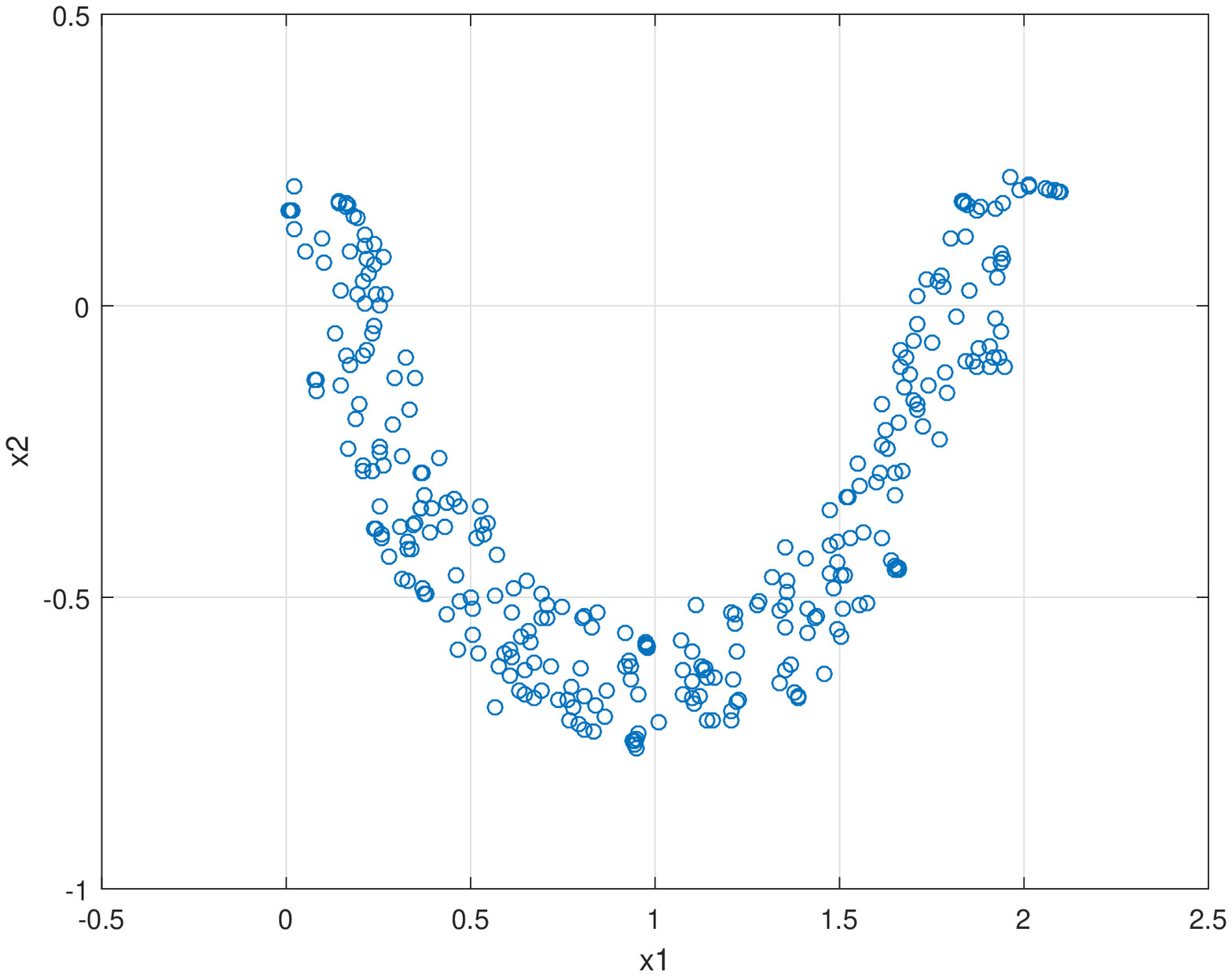} & \includegraphics[scale=0.3]{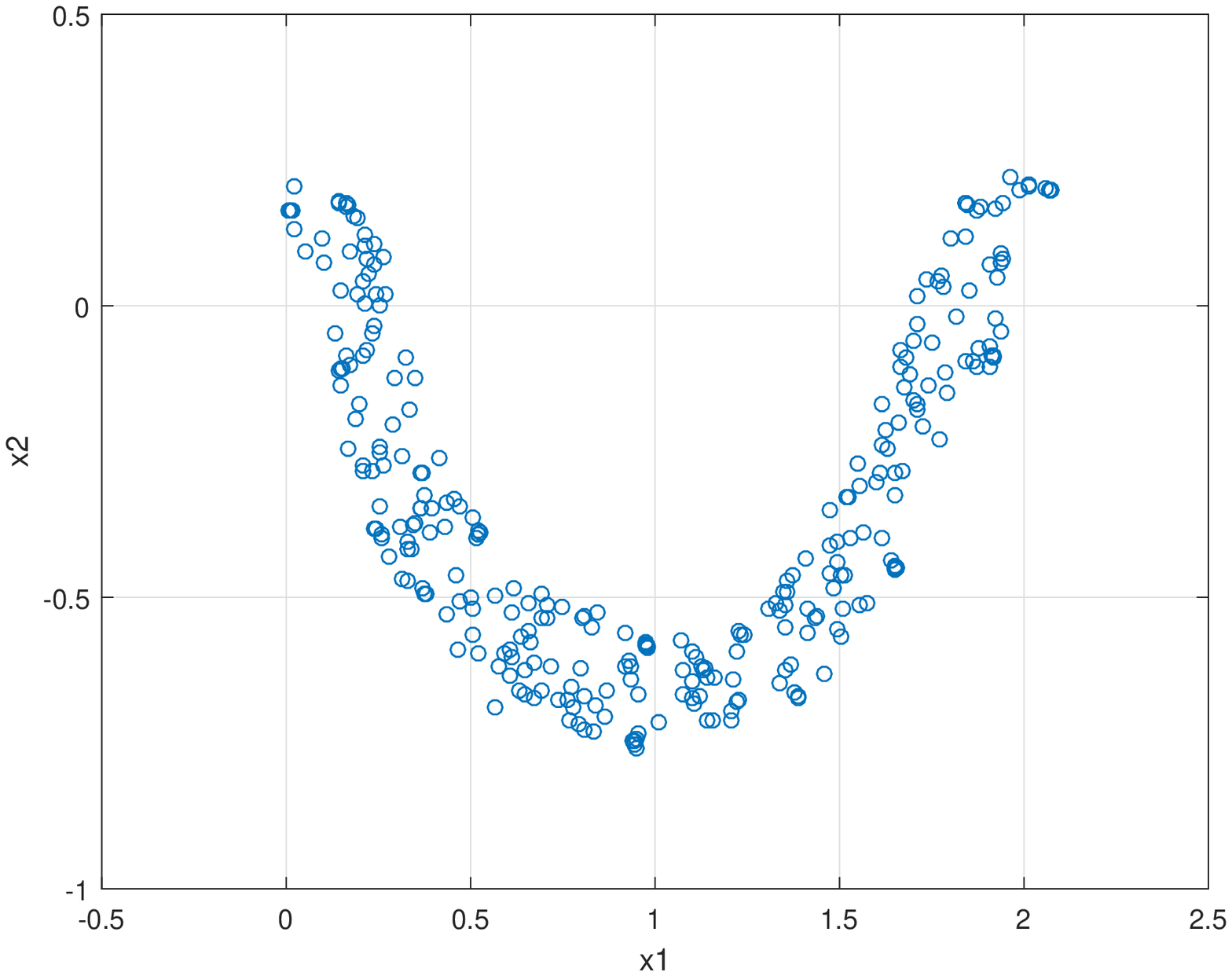}\\
c) Iteration $\#4$ & d) Iteration $\#6$ \\
\\
\multicolumn{2}{c}{\textbf{Fig. 5 } Performance of the proposed method for non-linear distribution in the presnece of $15\%$ outliers}

\end{tabular}
\end{center}

\end{table*}

\subsection{Real world data set}
In order to have a more realistic evaluation of the proposed methodm it should be tested on real world data set. Pen digit data set as of one UCI data sets is choosed for the evaluation \cite{UCI}. 
\subsubsection{Evaluation metric}
The goal of a denoising method is to recover the real distribution of data from the noisy one. Therefore the resulting distribution should close to the real distribution. In order to measure the diffrence between the resulting and real distributions, Divergence is employed. Divergence distance measures the similarity of two probability distributions \cite{Theod}.

\begin{equation}
 D_{pq}=E\{ln \frac{p(x)}{q(x)}\}=\int p(x) ln \frac{p(x)}{q(x)} d\textbf{x} 
\end{equation}
similar discussion holds for class $\omega_2$

\begin{equation}
 D_{qp}=E\{ln \frac{q(x)}{p(x)}\}=\int q(x) ln \frac{q(x)}{p(x)} d\textbf{x} 
\end{equation}

The sum 

\begin{equation}
d=D_{pq}+D_{qp}
\end{equation}
is called divergence and can be used as a discriminatory measure for the distributions $p$ and $q$.\\

Assuming that the density functions are Gaussians $\mathcal{N}(\mu_p,\Sigma_p)$ and $\mathcal{N}(\mu_q,\Sigma_q)$, the divergence can be computed as:\\

\begin{center}
$d_{pq}=\frac{1}{2}trace\{ \Sigma^{-1}_p \Sigma_q+\Sigma^{-1}_q \Sigma_p-2I\}$

\end{center}
\begin{equation}
+\frac{1}{2}(\mu_p-\mu_q)^T(\Sigma^{-1}_p+\Sigma^{-1}_q)(\mu_p-\mu_q)
\end{equation}

By the above definitions, the Divergence value for the real distribution and denoised distribution should be a minimum as possible for a good denoising method. The values of Divergence between the two distributions for different rates of outliers are summarized in Table I.  As it can be seen, the proposed method yeilds distribution for which the Diverence value between them and the real distribution of the sample is small. \\

In Fig. 6 an illustration of the performance of the proposed method on Pen digits data set is provided. Fig. 6.a) illustrates the Pen digits data set in 3 dimension. Fig. 6.b) show the data set with $10\%$ outliers and in Fig. 6.c) the data set is shown after outlier absorbing. 

\begin{table*}[ht]
\begin{center}
\begin{tabular}{cc}
\includegraphics[scale=0.5]{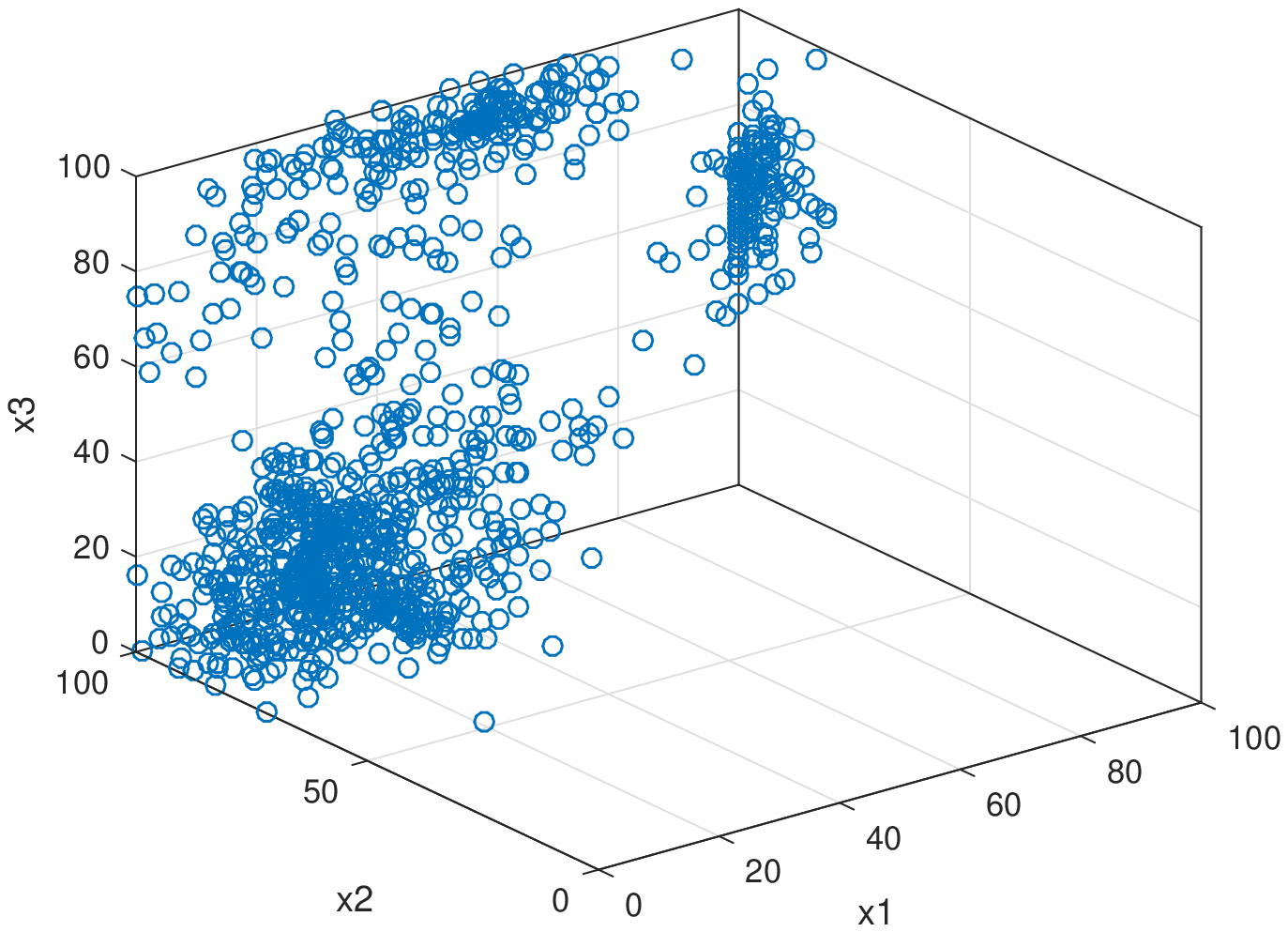}&\includegraphics[scale=0.5]{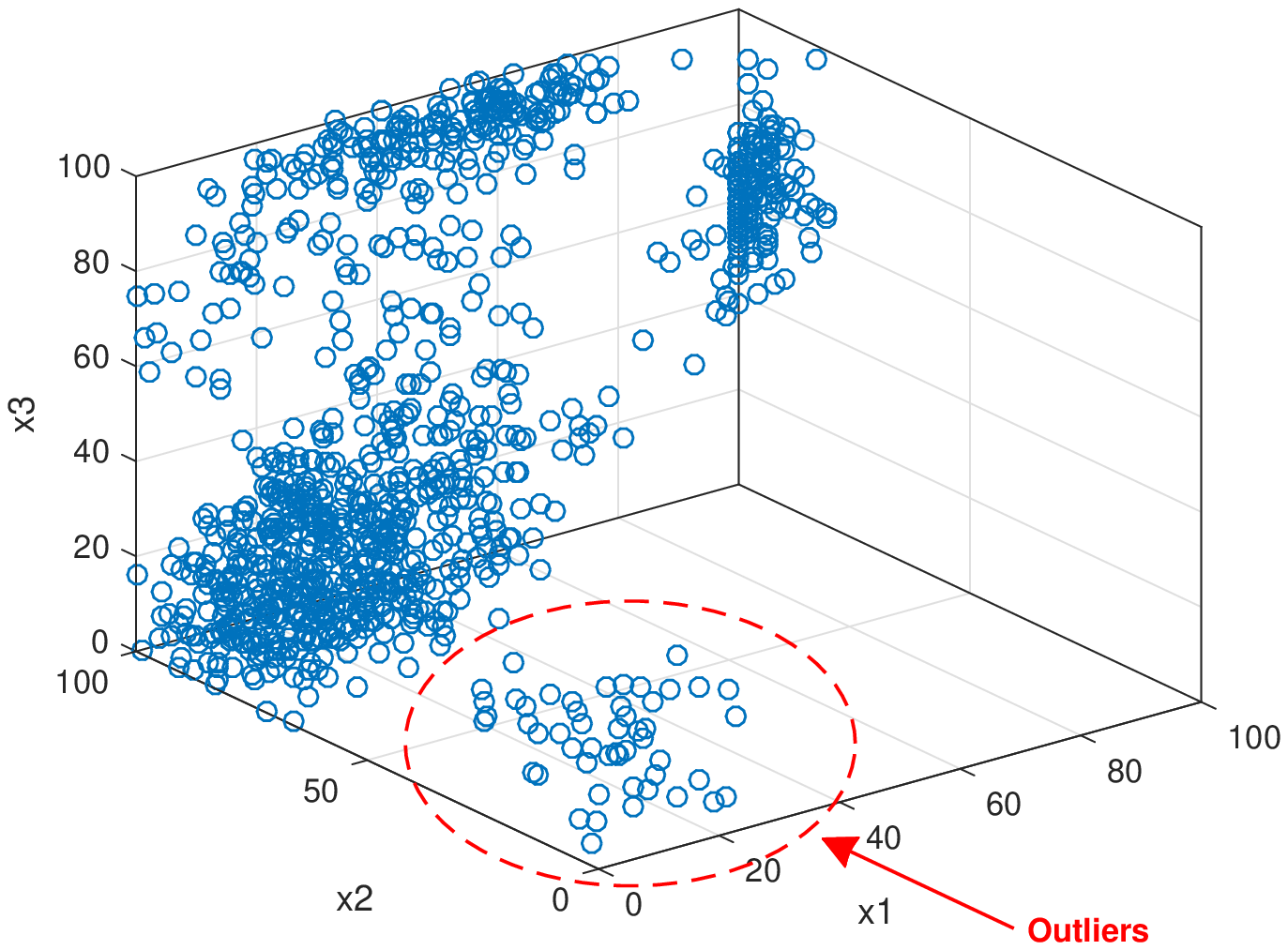}\\
a) An illustration of Pen digits data set in 3 dimension & b) Pen digits data with outliers\\
\multicolumn{2}{c}{\includegraphics[scale=0.5]{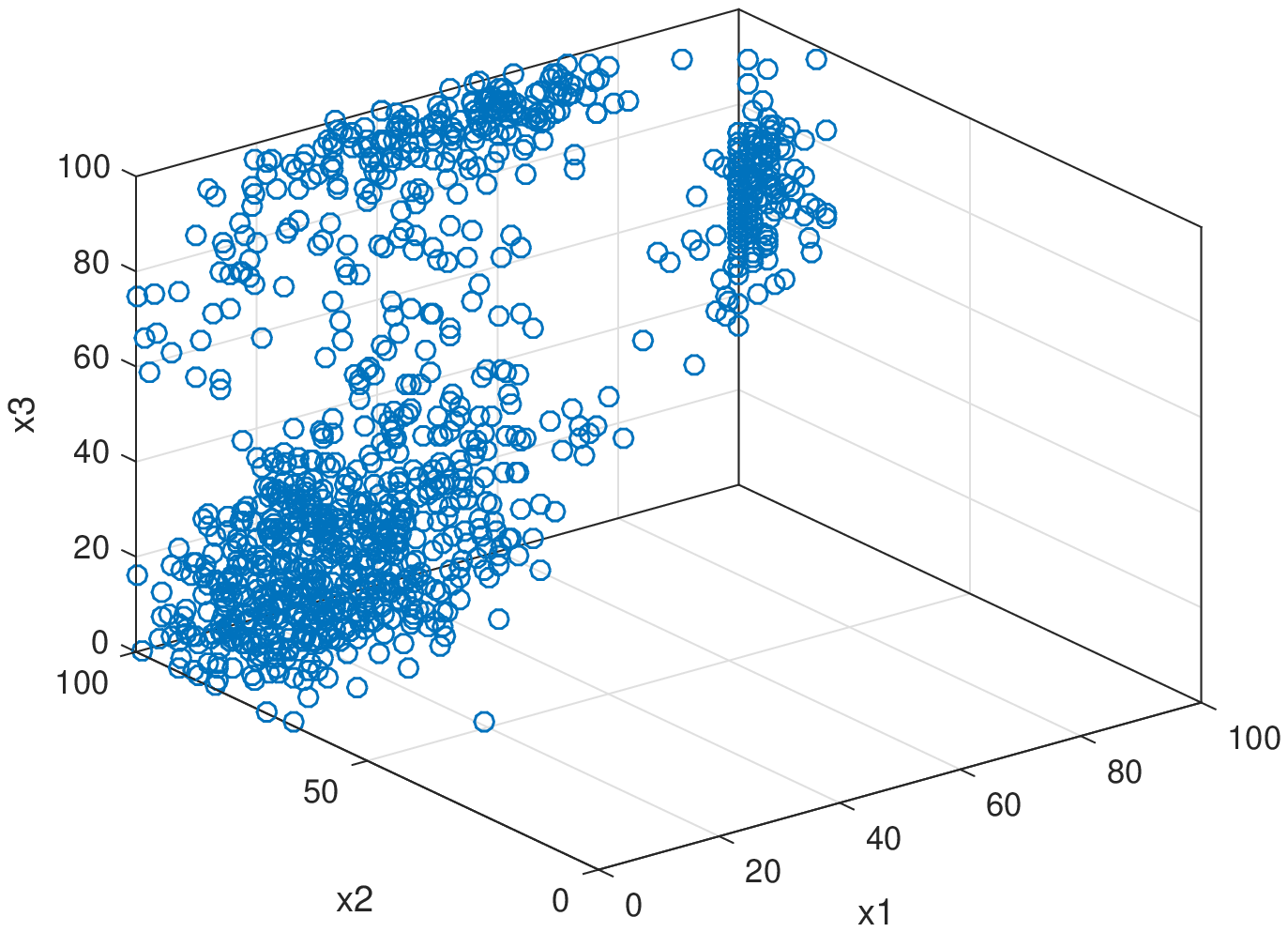}}\\
\multicolumn{2}{c}{c) Pen digits data set after outlier absorbing}\\
\\
\multicolumn{2}{c}{\textbf{Fig. 6} Performance of the proposed method on Pen digits data set}
\end{tabular}
\end{center}

\end{table*}


\section{Conclusion}
In this paper a new method for dealing with outliers was proposed. The poropsed method employs the local and global information of instance to overcome the outlier problem. As the experimental results showed, the combination leads to a more robust method for dealing with outliers. For future work we plan to extend our work to multi-class classification case.

\begin{table}[ht]
\centering
\caption{Outlier aobsoring results for Pen digits data set}
\label{my-label}
\begin{tabular}{ccc}
\hline
Outlier rate & Div before absorbing & Div after absorbing \\
\hline
1\%          & 1.1169               & \textbf{0.1508}     \\
5\%          & 7.2271               & \textbf{0.4685}     \\
10\%         & 15.7836              & \textbf{1.1856}     \\
15\%         & 24.1454              & \textbf{2.6886}  \\
\hline\\  
\end{tabular}
\end{table}



%

\end{document}